\documentclass{article}

\usepackage{microtype}
\usepackage{graphicx}
\usepackage{subfigure}
\usepackage{booktabs} % for professional tables

\usepackage{hyperref}

\usepackage[accepted]{icml2025}

\usepackage{amsmath}
\usepackage{amssymb}
\usepackage{mathtools}
\usepackage{amsthm}

\usepackage[capitalize,noabbrev]{cleveref}

\theoremstyle{plain}
\newtheorem{theorem}{Theorem}[section]
\newtheorem{proposition}[theorem]{Proposition}
\newtheorem{lemma}[theorem]{Lemma}
\newtheorem{corollary}[theorem]{Corollary}
\theoremstyle{definition}
\newtheorem{definition}[theorem]{Definition}
\newtheorem{assumption}[theorem]{Assumption}
\theoremstyle{remark}

\usepackage[textsize=tiny]{todonotes}

\DeclareMathOperator*{\argminA}{arg\,min}
\usepackage[overload]{empheq}

\usepackage{multirow}
\usepackage{booktabs,arydshln}
\makeatletter
\def\adl@drawiv#1#2#3{%
        \hskip.5\tabcolsep
        \xleaders#3{#2.5\@tempdimb #1{1}#2.5\@tempdimb}%
                #2\z@ plus1fil minus1fil\relax
        \hskip.5\tabcolsep}
\newcommand{\cdashlinelr}[1]{%
  \noalign{\vskip\aboverulesep
           \global\let\@dashdrawstore\adl@draw
           \global\let\adl@draw\adl@drawiv}
  \cdashline{#1}
  \noalign{\global\let\adl@draw\@dashdrawstore
           \vskip\belowrulesep}}
\makeatother

\icmltitlerunning{Energy-Based Preference Model Offers Better Offline Alignment}

\begin{document}

\twocolumn[
\icmltitle{Energy-Based Preference Model Offers Better Offline Alignment than the Bradley-Terry Preference Model}



\icmlsetsymbol{equal}{*}

\begin{icmlauthorlist}
\icmlauthor{Yuzhong Hong}{equal}
\icmlauthor{Hanshan Zhang}{equal}
\icmlauthor{Junwei Bao}{}
\icmlauthor{Hongfei Jiang}{}
\icmlauthor{Yang Song}{}
% \icmlauthor{Yuzhong Hong}{equal,yyy}
% \icmlauthor{Hanshan Zhang}{equal,yyy,comp}
% \icmlauthor{Firstname3 Lastname3}{comp}
\end{icmlauthorlist}


\icmlcorrespondingauthor{Yuzhong Hong}{first1.last1@xxx.edu}
\icmlcorrespondingauthor{Hanshan Zhang}{first2.last2@www.uk}

\icmlkeywords{Machine Learning, ICML}

\vskip 0.3in
]



\printAffiliationsAndNotice{\icmlEqualContribution} % otherwise use the standard text.

\begin{abstract}
Since the debut of DPO, it has been shown that aligning a target LLM with human preferences via the KL-constrained RLHF loss is mathematically equivalent to a special kind of reward modeling task. Concretely, the task requires: 1) using the target LLM to parameterize the reward model, and 2) tuning the reward model so that it has a 1:1 \emph{linear relationship} with the true reward. However, we identify a significant issue: the DPO loss might have multiple minimizers, of which only one satisfies the required linearity condition. The problem arises from a well-known issue of the underlying Bradley-Terry preference model: it does \emph{not always} have a unique maximum likelihood estimator (MLE). Consequently, \textbf{the minimizer of the RLHF loss might be unattainable because it is merely one among many minimizers of the DPO loss}. As a better alternative, we propose an energy-based model (EBM) that \emph{always} has a unique MLE, inherently satisfying the linearity requirement. To approximate the MLE in practice, we propose a contrastive loss named \textbf{E}nergy \textbf{P}reference \textbf{A}lignment (EPA), wherein each positive sample is contrasted against one or more strong negatives as well as many \emph{free} weak negatives. Theoretical properties of our EBM enable the approximation error of EPA to \emph{almost surely} vanish when a sufficient number of negatives are used. Empirically, we demonstrate that EPA consistently delivers better performance on open benchmarks compared to DPO, thereby showing the superiority of our EBM.
\end{abstract}
\begin{figure}
  \centering
  \includegraphics[width=0.41\textwidth]{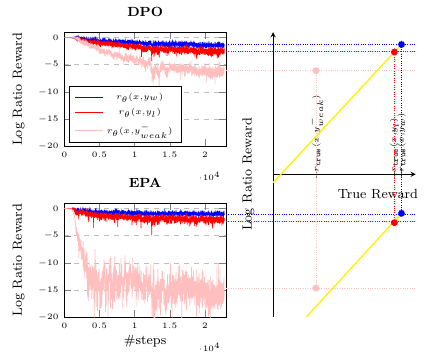}
  \caption{
  %Very weak negatives such as mismatched inputs and outputs are almost impossible to be preferred by humans, i.e., $r_{\text{true}}(x, y_w) > r_{\text{true}}(x, y_l)\gg r_{\text{true}}(x, y_{weak}^{-})$. Therefore, a necessary condition for the learned log ratio reward to be linearly related with the true reward is to expect that $r_{\theta}(x, y_w) > r_{\theta}(x, y_l)\gg r_{\theta}(x, y_{weak}^{-})$. However, 
  Samples are off from the slope-1 linearity (yellow lines) after training with DPO. 
  Given an extremely undesirable $y_{weak}^{-}$ (i.e., it has very small $r_{\text{true}}$), 
  %the linearity
  %, equivalent to the minimizer of the RLHF loss
  %is attainable only if 
  its $r_{\theta}$ has to be as sufficiently small as $r_{\text{true}}$ to attain the linearity.
  }
  \label{fig:motivation}
\end{figure}
\begin{figure*}
    %\centering
    % \hspace*{-1.1cm}\includegraphics[width=1.1\textwidth]
    \includegraphics[width=1.03\textwidth]{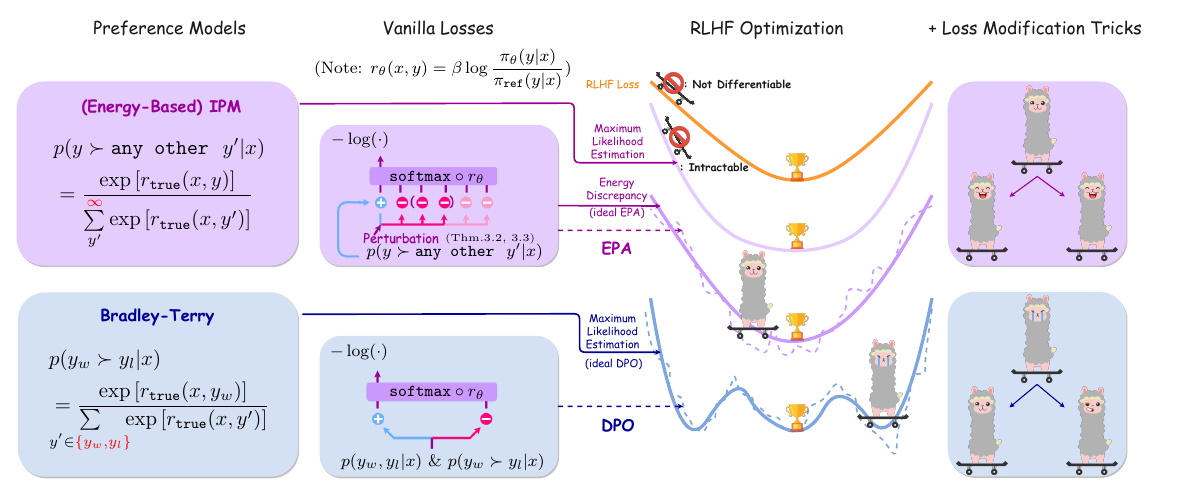}
    \caption{An illustration of the contributions of the paper. Our core argument is that an Energy-Based model (EBM) is a better alternative to the Bradley-Terry model (BTM) due to its guaranteed unique existence of \emph{maximum likelihood estimator} (MLE) (which is identical to the minimizer of the RLHF loss). The advantage of our EBM comes from its intrinsic consideration of the infinity in the size of the space of $y|x$, whereas BTM ignores issues caused by the pair sampling distribution ($p(y_w,y_l|x)$) in such infinite space. Hence we name our EBM the \emph{Infinite Preference Model}. Although approximating the MLE with our proposed EPA loss introduces inevitable error in practice, we find that it is still empirically better performing than its counterpart -- DPO, with or without loss modification techniques presented in previous offline alignment literatures.}
    \label{fig:illustration}
\end{figure*}
\section{Introduction}
\label{sec:intro}
Reinforcement Learning with Human Feedback (RLHF) \citep{christiano2017deep} has been widely used to align a large language model (LLM) with human preference. The canonical RLHF objective \citep{ziegler2019fine,stiennon2020learning,ouyang2022training,perez2022red} is defined as follows (given $x$):
\begin{equation}
  \begin{aligned}
    & \mathcal{L}_{\text{RLHF}} = \\
    &-\mathop{\mathbb{E}}_{\pi_{\theta}(y|x)}[r_{\text{true}}(x, y)] + \beta \text{KL}[\pi_{\theta}(y|x)||\pi_{\text{ref}}(y|x)]
  \end{aligned}
\end{equation}
where $\pi_{\theta}(y|x)$ is the target LLM (i.e., the policy) to tune, $\pi_{\text{ref}}(y|x)$ a frozen LLM initialized identically as the target LLM and $r_{\text{true}}(x, y)$ a reward to maximize.

The $\mathcal{L}_{\text{RLHF}}$ as defined above is not differentiable w.r.t $\theta$ \citep{ziegler2019fine,rafailov2023direct}, hence not SGD-friendly. 
%While classical approach is to use online RL methods such as PPO, there is also RL-free methods different approach by converting the optimization problem into a differentiable equivalent. 
Luckily, it has been shown that the unique minimizer of $\mathcal{L}_{\text{RLHF}}$ can be analytically expressed \cite{korbak2022reinforcement}. Then, \citet{rafailov2023direct} further reformulate the analytical minimizer as the unique solution to the following set of equations:
\begin{align}[left=\empheqlbrace]
    \text{\small$r_\theta(x, y)$} &\text{\small$= \beta \log \frac{\pi_{\theta}(y|x)}{\pi_{\text{ref}}(y|x)}$}\\
    \text{\small$r_\theta(x, y)$} &\text{\small$= r_{\text{true}}(x, y) + C(x)$}
\end{align}
Eq.(2) defines $r_{\theta}$ as the \emph{log ratio reward} and Eq.(3) states that there holds a \emph{slope-1 linearity} between the log ratio reward and the true reward. This formulation implies that as long as we can find a differentiable objective function $\mathcal{L}(r_{\theta})$ to achieve Eq.(3) and $r_{\theta}$ is parameterized by Eq.(2), we can convert the RLHF problem into an offline supervised task. This is the approach of interest in this paper, a drastically different one from classical online RL methods such as PPO.

\subsection{Background \& Motivation}

The poster child example of this offline approach is DPO \cite{rafailov2023direct}:
\begin{equation}
  \begin{aligned}
    & \text{\small$\mathcal{L}_{\text{DPO}}^{\text{ideal}}(r_{\theta}) =$} \\
    &\text{\small$-\mathop{\mathbb{E}}_{p(y_w,y_l|x)}\mathop{\mathbb{E}}_{p(y_w\succ y_l|x)}[\log \sigma (r_\theta(x, y_w) - r_\theta(x, y_l))]$}
  \end{aligned}
\end{equation}
where $y_w$ and $y_l$ are two responses and $y_w\succ y_l$ means $y_w$ is prefered to $y_l$ given the prompt $x$. The ideal\footnote{``ideal" in the sense that the expectations in its loss function are accurately computed. In practice, they can only be approximated, which can introduce an additional error.} DPO loss is essentially the maximum likelihood estimation loss of the Bradley-Terry model (BTM) who posits a sigmoidal relationship between $p(y_w\succ y_l|x)$ and $r_{\text{true}}(x,y_w) - r_{\text{true}}(x,y_l)$.
%famous in the rank-learning literature \citep{ford1957,simons1999asymptotics,han2020asymptotic,pmlr-v119-hendrickx20a,pmlr-v162-bong22a,wu2022asymptotic}. 
\emph{If the maximum likelihood estimator (MLE) uniquely exists}, \citet{rafailov2023direct} show that the MLE will make the slope-1 linearity hold. 

However, as alluded to by our framing, we argue that \textbf{it is \emph{false} to conclude that the slope-1 linearity (i.e., the minimizer of the RLHF loss) is guaranteed to be reached with DPO}.
%, which is unfortunately echoed by many follow-up researches \citep{azar2023general,ethayarajh2024kto}. 
%The root cause of the falsehood is that 
The reason is that the unique existence of BTM's MLE (i.e., the minimizer of $\mathcal{L}_{\text{DPO}}$) is not guaranteed without some non-trivial constraints on the structure of $p(y_w,y_l|x)$, a well-known issue of BTM given an infinite candidate space (i.e., that of $y|x$) in the literature on \emph{learning to rank}
%which is a data distribution independent from BTM.
%However, many of these assumptions are too strong to meet in the case of RLHF of LLM due to the infinite space of $y|x$ 
\citep{ford1957,simons1999asymptotics,han2020asymptotic,pmlr-v119-hendrickx20a,pmlr-v162-bong22a,wu2022asymptotic}.
%For example,
%\footnote{We discuss other necessary assumptions that cannot be met in the Appendix X. However, a single one is enough to show the insufficiency of minimizing $\mathcal{L}_{\text{DPO}}$ to achieve the slope-1 linearity.},
%one of the most basic necessary assumptions is that the comparison graph built by the pair samples from $p(y_w,y_l|x)$ should be a \emph{connected} graph with nodes covering all the candidates. Since there are an infinite number of candidates of $y$ given any $x$, this assumption cannot be met with a finite amount of samples of $y$\footnote{Even if we could sample (but we \emph{cannot}) an arbitrarily large number of unique candidates (denoted as $d$), the loosest sufficient condition proven so far requires the number of sampled comparisons to increase no slower than $O(\log d)$ \citep{pmlr-v162-bong22a}. Let alone other necessary assumptions that cannot be met as we discuss in Appendix X.}. 
It is also related to the theoretical issues around \emph{dataset coverage} in the RL literature \cite{kakade2002approximately,munos2008finite,zhan2022offline}. Moreover, \citet{tang2024generalized} have shown that when offline data are from $\pi_{\text{ref}}$ (a usual practice), any pair-wise loss will cease to correlate with $\mathcal{L}_{\text{RLHF}}$ when $\pi_{\theta}$ deviates enough from $\pi{_{\text{ref}}}$ due to reward maximization. This is the theoretical side of our motivation.

A preliminary experiment also evidently shows that DPO does not achieve the slope-1 linearity. Assuming it does, we should expect $r_{\theta}(x, y_w) > r_{\theta}(x, y_l)\gg r_{\theta}(x, y_{weak}^{-})$ because the true rewards are certainly so if \emph{weak} negatives ($y_{weak}^{-}$) are mismatched inputs and outputs as they are almost impossible to be preferred. 
%One of the most intuitive ways to sample \emph{weak-enough} negatives ($y_{weak}^{-}$) is simply mismatching inputs and outputs in the training data. 
However, as shown in Figure \ref{fig:motivation}, we find that their log ratio rewards are not substantially lower than the strong negatives ($y_l$) when training with DPO. This empirical phenomenon motivates us to find a method that utilizes the \emph{free} signal offered by $y_{weak}^{-}$.

\subsection{Contributions}
Our proposal brings the theoretical and the empirical sides together. 
%However, Instead of modeling other distributions as for KTO and NCA, we still model the preference distribution. 
As shown in Figure \ref{fig:illustration}, we argue that an Energy-Based model (EBM) called the \emph{Infinite Preference Model} (IPM) is superior to BTM in preference modeling for offline alignment based on the following contributions:
%because it has a better theoretical property: guaranteed unique existence of MLE which is mathematically equivalent to slope-1 linearity. We summarize our contributions as follows:
\begin{itemize}
    \item theoretically showing IPM has guaranteed unique existence of its MLE, equivalent to the minimizer of the RLHF loss;
    \item the proposal of EPA, an offline contrastive loss to estimate the MLE of IPM by explicitly using weak negatives in addition to strong negatives;
    \item a new state of the art of offline alignment on open benchmarks when given the same settings of training data and usage of tricks to tweak the losses.
\end{itemize}

\section{Related work}
\subsection{DPO and its recent improvements}
The first approach to avoid DPO's theoretical issue is to use non-BTMs to model data distributions.
%in the sense that they all use the log ratio reward to parameterize a reward model. They only differ in their loss functions to reach the minimizer of $\mathcal{L}_{\text{RLHF}}$.
\citet{rafailov2023direct} suggest the DPO's counterpart for the Plackett-Luce Model (we refer to it as DPO-PL), which is a generalized version of BTM for K-wise comparison.
IPO \citep{azar2023general} uses a different pair-wise preference model than BTM. The loss derived from that model can be interpreted as: the difference of log ratio rewards of the $y_w$ and $y_l$ regresses to a constant. However, \citet{tang2024generalized} show that IPO is still incapable of optimizing $\mathcal{L}_{\text{RLHF}}$, similar to DPO.
\citet{ethayarajh2024kto} (KTO) point out some limitations of modeling human preference with a pair-wise model. Instead, they independently model a data distribution for desirable samples and another one for undesirable samples. However, such data distributions do not reflect how most benchmark datasets are sampled. This could be the reason why some empirically driven studies find that KTO underperforms DPO on these benchmarks \cite{meng2024simpo,zhou2024wpo}.

The second approach is to tweak the DPO loss. Some loss-tweaking tricks can be effective on their own. For example, cDPO \citep{eric2023cdpo} uses label smoothing to alleviate DPO's overfitting problem. \citet{park2024disentangling} (R-DPO) introduce a length penalty on the log ratio reward to make DPO less prone to the verbosity bias. \citet{amini2024direct} (ODPO) add a dynamic margin between $y_w$ and $y_l$ based on the intuition that some pairs have stronger or weaker desirability gaps than others. The most effective one discovered so far is on-policy weighting (WPO) \cite{zhou2024wpo}. Its idea is to approximate the on-policy training scenario by assigning larger weights to the loss of samples closer the current policy at each step and smaller weights to that of less closer ones. 
Other tricks come in combinations. For example, CPO \cite{xu2024contrastive} removes the reference model in the log ratio reward and add an SFT loss at the same time. ORPO \cite{hong2024orpo} is an improvement over CPO by adding yet another set of tricks: normalizing the policy to the token level (length normalization) and then contrasting the policy distribution with one minus itself. 
%PRO \cite{song2024preference} is another example of a complex combination of tricks. On top of the tricks of CPO, it not only requires length normalization but also necessitates more responses per prompt and an additional reward model to score every response. 
To separate the wheat from the chaff, \citet{meng2024simpo} find the most simple and effective recipe: removing the reference model, adding a margin and applying length normalization, which gives rise to SimPO. 

There is also a hybrid approach of the above two: non-BTMs + tricks. For example, PRO \cite{song2024preference} is on top of DPO-PL, and BCO \cite{jung2024binary} on top of KTO.

The problem with applying tricks is that there is usually a lack of theoretical justification on how they are related to the minimizer of the RLHF loss.

% \subsection{Improvements of $\mathcal{L}_{\text{RLHF}}$}
% Another major branch of RLHF research focuses on modifying the canonical $\mathcal{L}_{\text{RLHF}}$ itself. Therefore, this approach is not directly comparable with ours in the narrow sense of optimizing the canonical $\mathcal{L}_{\text{RLHF}}$. However, in a broader sense of RLHF, the approach is comparable with ours in terms of the end performance of alignment with human preference.

\subsection{Fitting discrete EBMs}
To provide a theoretical background for our proposal, we give a concise review of the most related work on fitting discrete EBMs.

Energy-based models (EBM) \cite{lecun2006tutorial} are generative models that posit a Boltzmann distribution of data, i.e., $p(x) \propto \exp{(-E(x))}$ where $E(x)$ (called the \emph{energy function}) is a real-valued function to learn. An EBM is called discrete when the data point $x$ is defined on a discrete space. To fit $p(x)$ with maximum likelihood estimation requires the computation of the normalizer $\Sigma_{x^{'}}^{\infty}\exp{(-E(x^{'}))}$ (called the \emph{partition function}), which is intractable. Therefore, EBMs are usually learned with a tractable approximation.

The classical approach is to approximate the gradient of maximum likelihood estimation by online sampling from parameterized $p_{\theta}(x)$ with MCMC \cite{song2021train}. Although they are ideally effective, it is usually difficult or expensive to do such sampling, which harms practical results.
Therefore, there are also many MCMC-free methods \citep{meng2022concrete,hyvarinen2007some,dai2020learning,lazaro2021perturb,eikema2022approximate}. Recently, \citet{schroder2023training} have introduced the notion of energy discrepancy, whose unique global minimizer is identical to the MLE of the EMB in question. Hence, to find the MLE, one can simply minimize the energy discrepancy, which is feasible with SGD on offline data. For its simplicity, we derive EPA based on their theoretical framework.

\subsection{EBMs for RLHF}
EBMs are not rare in the RLHF research. One of the research directions is to formulate RLHF as a Distribution Matching problem:  minimization of the KL divergence between a target EBM that reflects human preference and the policy. The typical example for this approach is Distributional Policy Gradients (DPG) \cite{parshakovadistributional,khalifa2021a}. However, we would like to point out that our EBM is different and used for a different purpose. The EBM in DPG is a non-parametric one predefined as the learning signal. Our EBM is a parametric one to fit the distribution of data. The only connection between the two EBMs is that they are used to find the same optimal policy \cite{korbak2022reinforcement}.

\citet{Deng2020Residual} uses an EBM for language modeling. Their work essentially solves the self-play-like RLHF problem \citep{chen2024self}. They use the algorithm of Noise Contrastive Estimation (NCE) of fit their EBM. Although their EBM is also parametric, it fits the optimal policy distribution. Our EBM instead fits the preference distribution.
%Interestingly, the two EBMs also find the same minimizer (see Appendix x). 

\citet{chen2024noise} proposes two methods -- infoNCA and NCA, based on the same EBM as that of \citet{Deng2020Residual}. The NCA loss follows the same derivation of the loss proposed by \citet{Deng2020Residual} except that they parameterize their energy function differently. The infoNCA loss exhibits similarity to our loss. However, we will show that infoNCA is just a worse-performing ablation version of EPA.

\section{IPM: Our EBM for Preference Modelling}
\label{sec:ipm}
In the first subsection, we show that an energy-based model (EBM) is guaranteed to have a unique MLE which is equivalent to the minimizer of the RLHF objective. In the second subsection, based on a framework by \citet{schroder2023training}, we describe a general strategy to approximate the MLE using offline data. Using it will provide the theoretical account for our proposal in section \ref{sec:EPA}.
\subsection{Theoretical guarantee}
Given any $x$, it is obvious that the space of $y|x$ is infinitely large because $y$ can be any token sequence of unlimited length no matter how likely or unlikely it is a response to $x$. This infinity is problematic for BTM. For example, if there is a single $y$ that is never sampled, it is easy to refute the unique existence of BTM's MLE (see Proposition \ref{prop:btflaw}). However, EBM can naturally take the infinity into account to avoid the issue.
Specifically, we model a one-to-infinite preference (v.s. BTM and the more general Plakett-Luce model only model a one-to-finite-number preference) as follows:
\begin{equation}
  \begin{aligned}
    p(y|x)\coloneqq p(\forall y'\neq y:\ y\succ y'|x)
  \end{aligned}
\end{equation}
Namely, $p(y|x)$\footnote{One should not confuse $p(y|x)$ with $\pi(y|x)$ although both of them are distributions over $y$ given $x$. $p(y|x)$ is how likely humans would rate a $y$ as the best whereas $\pi(y|x)$ measures how likely $y$ is to be generated.} is the probability that candidate $y$ is preferred over all other candidates.
Under mild assumptions (Assumptions \ref{assume1} and \ref{assume2}) that make an EBM applicable, we define the Infinite Preference Model (IPM) to be the one that posits that $p(y|x)$ is a Boltzmann distribution induced by the corresponding true reward (i.e., using $-r_{\text{true}}(x,y)$ as the energy function):
%$$p(y|x) = \frac{\exp [r_{\text{true}}(x,y)]}{\Sigma_{y^{'}}^{\infty}\exp [r_{\text{true}}(x,y^{'})]}$$
\begin{equation}
  \begin{aligned}
    p(y|x) = \frac{\exp [r_{\text{true}}(x,y)]}{\Sigma_{y^{'}}^{\infty}\exp [r_{\text{true}}(x,y^{'})]}
  \end{aligned}
\end{equation}
%$where the denominator exists because of the Assumptions \ref{assume1} and \ref{assume2}.$
%This is essentially an EBM defined on an infinite discrete space with $-r(x,y)$ as its energy function.
%When $q_{\theta}(y|x)$ is parameterized with the log ratio reward and the above objective is minimized w.r.t $\theta$, the following theorem allows us to recover the absolute scoring of every possible $y$ up to a constant (i.e., the true reward $r(x, y)$ plus a unique $C(x)$) 
IPM is a better alternative to BTM because of the following theorem (see Appendix \ref{apx:ipm} for proof).

% Although the introduction of infinity seems pointless since it is practically infeasible, our extension is non-trivial. The significance comes from the following theorem (see Appendix \ref{apdx:b} for proof):
\begin{theorem}
\label{theorem:main}
when we parameterize the IPM as follows, the unique existence of the IPM's MLE is guaranteed and it will be reached if and only if the slope-1 linearity (i.e., Eq.(3)) holds between the log ratio reward and the true reward. 
%$$q_{\theta}(y|x) = \frac{\exp [r_{\theta}(x,y)]}{\Sigma_{y^{'}}^{\infty}\exp [r_{\theta}(x,y^{'})]}$$
\begin{equation}
  \begin{aligned}
    q_{\theta}(y|x) = \frac{\exp [r_{\theta}(x,y)]}{\Sigma_{y^{'}}^{\infty}\exp [r_{\theta}(x,y^{'})]}
  \end{aligned}
\end{equation}
where $r_{\theta}$ is defined as in Eq.(2).
\end{theorem}
Therefore, as long as we can find the MLE of the IPM parameterized as so, we are guaranteed to reach the minimizer of $\mathcal{L}_{\text{RLHF}}$ since it is the unique solution to Eq.(2) and Eq.(3).

On a side note, the IPM has been previously introduced by other studies on RLHF for a different purpose: to theoretically equate the maximization of $-\mathcal{L}_{\text{RLHF}}$ to the variational inference of the optimal policy with $\pi_{\text{ref}}$ as the prior \citep{korbak2022rl,yang2024asymptotics}. However, to the best of our knowledge, we are the first one to introduce IPM not just as a theoretical toy, but as a tool (when parameterized by the log ratio reward) to do proper RL-free RLHF.

\subsection{Offline approximation of MLE}
Despite the powerfulness of IPM, finding its MLE is a non-trivial task. Directly finding it with the minimization of the negative log likelihood $-\log q_{\theta}(y|x)$ is intractable because of the infinity in the denominator.

There are good tractable approximations but usually with complex online training algorithms. For simplicity and scalability purposes, we choose to follow \citet{schroder2023training}, who provide a general strategy that finds the optimal EBM by simple SGD with offline training data. The strategy is based on two theorems formally adapted for our purpose as follows.
\begin{theorem}
\label{theorem:ed}
For any random variable $Z$ with the conditional variance $\text{Var}[Y|Z]$ being positive, the global unique minimizer $r^{*}(x, y)$ of the functional \emph{Energy Discrepancy} (ED) defined as follows is the optimal IPM (i.e., $p(y|x)\propto \exp{[r^{*}(x,y)]}$).
\begin{equation}
  \begin{aligned}
    & \text{ED}_{x,p(y|x),p(z|y)}[r]\coloneqq\\
    &\mathop{\mathbb{E}}_{p(y|x)}\mathop{\mathbb{E}}_{p(z|y)}[\log\Sigma_{y'}{p(z|y')\exp{[r(x,y') - r(x, y)]}}]
  \end{aligned}
\end{equation}
\end{theorem}
\begin{theorem}
\label{theorem:EPA}
For any random variable $Z$ whose backward and forward transition probabilities from $Y$ solve the equation $\Sigma_{y}p(z|y)f(y)=\Sigma_{y}p(y|z)f(y)$ for an arbitrary $f$, the estimation error of the following statistic estimate of $\text{ED}_{x,p(y|x),p(z|y)}[r]$ vanishes almost surely when $N\xrightarrow{}\infty$ and $M\xrightarrow{}\infty$.
\begin{equation}
\label{eq:edloss}
  \begin{aligned}
    & \mathcal{L}[\theta|x]\coloneqq\\
    &\frac{1}{N}\Sigma_{i}^{N}\log (1 + \Sigma_{j}^{M}{\exp{[r_{\theta}(x,y_{-}^{i,j}) - r_{\theta}(x, y^{i})]}})\\
    &-\log(M)
  \end{aligned}
\end{equation}
where $\{y^{i}\}^{N}$ are samples from $p(y|x)$, $\{y_{-}^{i,j}\}^{M}$ from $p(y|z_{0})$, and $z_{0}$ a single sample from $p(z|y^{i})$.
\end{theorem}
A one-sentence interpretation of the above theorems is: if we have a particular kind of negative sampling strategy by perturbing observed preferred samples, we will learn the optimal IPM by minimizing a contrastive loss between the negatives and observed positives. Therefore, when the IPM is parameterized by the log ratio reward, we will find the exact minimizer of $\mathcal{L}_{\text{RLHF}}$ with this loss function.

Note that the property of the negative sampling source $Z$ as described in Theorem \ref{theorem:EPA} is just a sufficient condition as opposed to a necessary one. This leaves room for empirical discovery of better negative sampling strategies. A rule of thumb as suggested by \citet{schroder2023training} is that \textbf{$Z$ has to be informative of $Y$ and of high conditional variance at the same time}. This provides the intuition of our proposal in section \ref{sec:EPA}.
\section{EPA: A Practical Approximation}
\label{sec:EPA}
To introduce our loss, we first write the ideal loss in Eq.(\ref{eq:edloss}) in an equivalent form by removing the constant $\log(M)$ and moving a minus sign out of the logarithm:
\begin{equation}
\label{eq:edlossvariation}
  \begin{aligned}
    & \Tilde{\mathcal{L}}[\theta|x]\coloneqq\\
    &\frac{1}{N}\Sigma_{i}^{N}-\log \frac{\exp[r_{\theta}(x, y^{i})]}{\exp[r_{\theta}(x, y^{i})] + \Sigma_{j}^{M}\exp[r_{\theta}(x,y_{-}^{i,j})]} 
  \end{aligned}
\end{equation}
Now we propose our loss function in this \emph{negative log softmax} form with a specific negative sampling strategy in mind. 
%\subsection{Formulations}
\subsection{Narrow definition}
For the most classical setup, we assume we only have access to pair-wise preference data. In this setting, our loss for each mini-batch of $B$ samples ($\{(x^i, y_w^i, y_l^i)\}^B$)
%, assuming $i\neq j\Leftrightarrow x^i\neq x^{j}$
is defined as follows:
\begin{equation}
  \begin{aligned}
    & \text{\small $\mathcal{L}_{\text{EPA}}=$}\\
    & \text{\small $\frac{1}{B}\sum\limits_{i}^{B}-\log\frac{\exp [r_{\theta}(x^i,y_w^i)]}{\sum\limits_{\mathclap{j\in\{i\}\cup\mathcal{I}_{\text{wk}}}}\big(\exp [r_{\theta}(x^i,y_w^j)] + \exp [r_{\theta}(x^i,y_l^j)]\big)}$}
  \end{aligned}\hspace{-1.2em}
\end{equation}
where $\mathcal{I}_{\text{wk}}$ is a non-empty random subset of $\{1,\ 2,\ \dots,\ i-1,\ i+1,\ \dots ,\ B\}$, introducing negative samples that are mismatched responses originally sampled for other prompts. Its size $|\mathcal{I}_{\text{wk}}|=N^{-}_{weak}/2\in(0, B-1]$ is a hyperparameter. Note that our loss without $\mathcal{I}_{\text{wk}}$ reduces to the DPO loss.
We justify our choice of positives and negatives in EPA as follows:
\begin{enumerate}
    \item \textbf{Why is $y_w$ a good approximation of a positive sample from $p(y|x)$?} For a $y_w$ in the dataset, it may not be the best $y$, but there is only a finite number of potentially possible better ones according to Assumption \ref{assume1}. Also, since we know it is preferred over $y_l$ and infinitely many other arbitrary token sequences, it is a good approximation of a $y$ that is preferred over all other samples up to a small error.
    \item \textbf{Why use both $y_l$ and mismatched responses as negatives?} As stated at the end of section \ref{sec:ipm}, we want to draw the negatives from a perturbation source that is both informative of the positives and of high variance at the same time. For the informativeness, we consider \emph{strong} negatives $y_l$ because they are semantically close to $y_w$. For the high variance, we consider \emph{weak} negatives such as mismatched responses. We will show the effectiveness of such choice with ablation experiments in section \ref{sec:exp}.
\end{enumerate}

\subsection{General definition}
Note that the number of strong negatives in Eq.(11) is limited to 1 because of the given pair-wise data. This is not ideal for the approximation of IPM's MLE because the number of negatives should be large enough to reduce the approximation error (Theorem \ref{theorem:EPA}). Therefore, in order not to limit the power of IPM by the classical pair-wise data setup, we generalize our definition of EPA to circumstances where we can have access to more strong negatives (i.e., each $y_w$ is accompanied by $\{y_{l_1},\ y_{l_2},\ \dots\}$ instead of just one $y_l$). This is practically feasible because we can always sample less desirable responses from some LLM.

Hence, we define our loss in a more general form as follows:
\begin{equation}
  \begin{aligned}
    & \mathcal{L}_{\text{EPA}}^{\text{general}}=\\
    &\text{\tiny$\frac{1}{B}\sum\limits_{i}^{B}-\log\frac{\exp [r_{\theta}(x^i,y_w^i)]}{\exp [r_{\theta}(x^i,y_w^i)] + \sum\limits_{\mathclap{k\in\mathcal{I}_{\text{st}}}}\exp [r_{\theta}(x^i,y_{l_{k}}^i)] + \sum\limits_{\mathclap{j\in\mathcal{I}_{\text{wk}}}}\exp [r_{\theta}(x^i,y_{*}^j)]}$}
  \end{aligned}\hspace{-1.2em}
\end{equation}
where $\mathcal{I}_{\text{st}}$ contains $N^{-}_{strong}$ indices of available strong negatives; $\mathcal{I}_{\text{wk}}$ contains $N^{-}_{weak}$ indices of weak negatives; $y_{*}^{j}$ can be either $y_w^{j}$ or some $y_{l_k}^{j}$ ($j\neq i$). The justification for the choice of the positives, strong and weak negatives is the same as that of the narrow definition.

\subsection{Gradient analysis}
Using the chain rule consecutively on the negative log softmax and the log ratio reward, one can easily find the gradient of the EPA loss (the general one) as follows:
\begin{equation}\hspace*{-0.5cm}
  \begin{aligned}
     \nabla_{\theta}\mathcal{L}_{\text{EPA}}^{\text{general}}&=\\ 
     \text{\footnotesize$-\frac{\beta}{B}\sum\limits_{i}^{B}\Big($}&\text{\footnotesize$\underbrace{\sum\limits_{k\in\mathcal{I}_{\text{st}}}s^{i}_{l_k} \big(\nabla_{\theta}\log\pi_\theta({y^i_{w}}|x^i)-\nabla_{\theta}\log\pi_{\theta}({y^i_{l_k}}|x^i)\big)}_{\text{strong contrast}}$}\\
    +&\text{\footnotesize$\underbrace{\sum\limits_{j\in\mathcal{I}_{\text{wk}}}s^j\big(\nabla_{\theta}\log\pi_\theta({y^i_{w}}|x^i)-\nabla_{\theta}\log\pi_{\theta}({y^{j}_{*}}|x^i)\big)}_{\text{weak contrast}}\Big)$}
  \end{aligned}\hspace{-1.2em}
\end{equation}
where $s^{i}_{l_k}$ and $s^{j}$ are the softmax-ed values of the strong negative log ratio rewards and the weak negative log ratio rewards, respectively. They control the magnitude of the strong and weak contrast. When there is no weak contrast, the gradient reduces to that of the DPO loss if there is only one strong negative. Therefore, one can interpret the weak contrast as a regularization term added to DPO to prevent $\theta$ from moving to a direction that undesirably increases the likelihood of weak negatives.
\section{Experiments}
\label{sec:exp}
\subsection{Experimental setup}
\subsubsection{Training Data}
We consider the dataset of Ultrafeedback \cite{cui2024ultrafeedback} (denoted as `UF-all') and a widely used pair-wise version of it \cite{tunstall2023zephyr} (UF-binarized). The two datasets are ideal for our purpose besides their popularity. Firstly, in UF-all, there are 4 responses sampled from multiple sources for each prompt. This will allow training with our general version of EPA which can utilize multiple strong negatives. Secondly, in both UF-all and UF-binarized, the positive sample $y_w$ for each $x$ is the best one out of the 4 responses. This is an arguably close approximation to our assumption that positives are sampled from $p(\forall y'\neq y:\ y\succ y'|x)$.
\subsubsection{Evaluation} 
\begin{figure*}
  \centering\includegraphics[width=0.85\textwidth]{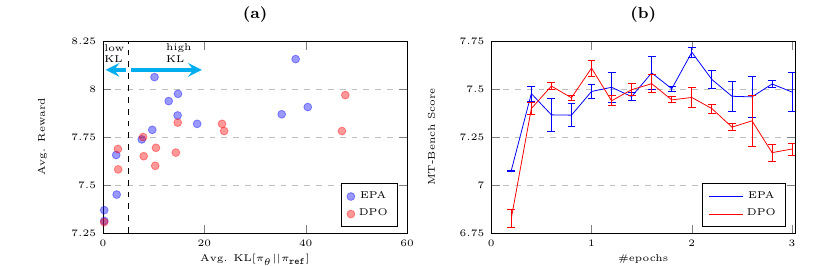}
  \caption{
  DPO vs. EPA (1:1:2) from the perspective of (a) KL-Reward frontier and (b) training dynamics.
  }
  \label{fig:kr_dynam}
\end{figure*}
Although Ultrafeedback is intended to reflect human preference, it is labeled by GPT-4 in reality. 
%Therefore, it makes sense to evaluate models trained on the two datasets with GPT-4 as well. 
Consequently, we consider MT-Bench \cite{zheng2024judging} which also uses GPT-4 to score a response on a scale of 1-10. The metric is the average score for 80 single-turn conversations and 80 multi-turn conversations. We also consider Alpaca-Eval 2.0 \cite{dubois2024length} because of its high correlation with human preference, the ultimate concern for RLHF. Its metrics are win-rates (with or without \emph{length control}) against GPT-4-turbo across 805 test samples with the judge being GPT-4-turbo itself. We report them in the format of ``length controlled win-rate / win-rate" in the experiment results.
%instead of a real-valued score for each response. Moreover, the judge used in Alpaca-Eval 2.0 is GPT-4-turbo rather than GPT-4. 
%However, we still consider it because it highly correlates with human preference, the ultimate concern for RLHF. 
For evaluation on Alpaca Eval 2.0, we use the default decoding parameters in the Huggingface implementation. For MT-Bench, we use the ones specifically required by the benchmark.

\subsubsection{Baselines \& loss modification tricks}
\begin{table}[ht!]
    \centering
    \begin{tabular}{cccc}
        \toprule
        \multicolumn{1}{c}{\textbf{Training Data}} & 
         \multicolumn{1}{c}{\textbf{Method}} & \multicolumn{1}{c}{\textbf{AE 2.0 (\%)}} & \multicolumn{1}{c}{\textbf{MT-Bench }} \\
        \midrule
        & SFT & 8.16 / 5.47 & 6.44 \\
       \cdashlinelr{1-4}
       \multirow{5}{*}{UF-binarized} & +DPO & 17.43 / 15.24 & 7.55 \\
        & +IPO & 12.97 / 10.13 & 7.31 \\
        &   +KTO & 12.62 / 11.29 & 7.21 \\
        &   +NCA & 14.64 / 11.27 & 7.39 \\
        \cdashlinelr{2-4}
        &   +EPA & \textbf{19.20} / \textbf{19.26} & \textbf{7.71} \\
            \midrule
        \multirow{4}{*}{UF-all} &   +DPO-PL & 15.95 / 14.68 & 7.57 \\
         &   +NCA & 15.08 / 11.85 & 7.28 \\
        &   +infoNCA & 17.30 / 16.25 & 7.50 \\
        \cdashlinelr{2-4}
        &    +EPA & \textbf{22.03} / \textbf{21.44} & \textbf{7.58} \\
           
        \bottomrule
    \end{tabular}
    \caption{EPA beats all other baselines either for pair-wise data or for data with $>2$ responses for each prompt. }
    \label{tab:alignment}
\end{table}
For fair comparison, we only consider methods from the approach that explicitly aims to minimize $\mathcal{L}_{\text{RLHF}}$ with specific probabilistic models about data distributions. Therefore, we consider DPO, IPO, KTO and NCA for the classical pair-wise data setting. We consider DPO-PL, NCA, and infoNCA for the general setting where there are multiple responses for each prompt in the dataset.

Loss modification tricks are not considered as baselines because they are \emph{orthogonal} to our proposal. Comparing BTM+tricks to EBM would be comparing apples to oranges. Instead, we consider applying the tricks to both EPA (the narrow one for fair comparison) and DPO to further verify our core argument about EBM's superiority over BTM. The tricks in consideration are those used in SimPO, R-DPO, CPO and WPO (Details in Table \ref{tab:listoftricks} in the Appendix \ref{apdx:b}).

\subsubsection{Implementation}
We use mistral-7b-sft-beta\footnote{huggingface.co/HuggingFaceH4/mistral-7b-sft-beta} as the reference model and for the initialization of policy in our paper. We train all models in this paper for 3 epochs with LoRA ($r=16$, $\alpha=16$, dropout $=0.05$). Whenever comparing among different methods, we pick the one out of the three checkpoints with the best MT-Bench score for each method. For fair comparison of baseline models, we fix $\beta$ to 0.01. It is more of a \emph{control variable} than a hyperparameter because it is a given component of the RLHF objective which all baselines are aimed to optimize.
We only vary $\beta$ for them when probing their KL-Reward frontiers. 
For comparison of loss modification tricks, since the RLHF objective is not necessarily the purpose, we use the best $\beta$ and other hyperparameters specific to each method as reported in previous work (e.g., the tricks used in SimPO are only competitive when $\beta=2.0$ for the Mistral model). Learning rate is grid-searched for each method among $\{1e-5, 5e-6, 1e-6\}$. 
%Unless explicitly stated otherwise, we report results of the narrowly defined EPA with $|\mathcal{I}|=1 (N_{weak}^-=2)$ trained on UF-binarized.

\subsection{Results and analysis}
\subsubsection{EPA performs better than baselines}
As shown in Table \ref{tab:alignment}, we can see EPA consistently achieves the highest scores and hence a new state of the art. Note that other baselines generally perform even less well than DPO. This makes BTM the strongest baseline for EBM. 
%As in Figure X, we can also see that EPA is better than other baselines in terms of Reward-KL frontier.

%Since our EBM and all the baselines essentially model rewards, we compare them on Reward Bench. As we can see, the log ratio reward trained with EPA is also better than that trained with DPO, a strong evidence that EBM is closer to achieving the linearity between $r_{\theta}$ and $r_{\text{true}}$.

\subsubsection{EPA $>=$ DPO for the optimization of $\mathcal{L}_{\text{RLHF}}$}
To compare our EBM with its most competitive baseline BTM in detail, we come back to the starting point -- optimizing $\mathcal{L}_{\text{RLHF}}$. We study from two perspectives of the optimization problem. Both perspectives involve multiple checkpoints beyond the single best one for each method (e.g., Table \ref{tab:alignment}), offering a more comprehensive comparison.

First, we study how well each method balances the KL term and the reward term in $\mathcal{L}_{\text{RLHF}}$ with varying $\beta\in\{0.01, 0.02, 0.03, 0.04,0.05,0.1,0.5\}$. Both terms are computed on the 80 single-turn prompts in the MT-Bench dataset. We estimate KL with 20 response samples per prompt from each policy distribution. We use the GPT-4 score produced by MT-Bench as an alias for the true reward. As shown in Figure \ref{fig:kr_dynam}.(a), in the high-KL region, EPA generally achieves higher reward than DPO. The two become indistinguishable only in the low-KL region.

Second, to understand how EPA differs from DPO in terms of the dynamics during the optimization process, we test the MT-Bench score of the checkpoint at every 20\% of an epoch. As shown in Figure \ref{fig:kr_dynam}.(b), EPA is less prone to overfitting and fits to the reward signal more steadily than DPO. The performance of DPO starts to degenerate rapidly after the first epoch. However, EPA reaches its peak performance at the end of the second epoch and overfits more slowly afterward. This is consistent with our gradient analysis in Section \ref{sec:EPA} that EPA is more regularized than DPO.

\subsubsection{Combining strong and weak negatives is effective}
\begin{table}
    \centering
    \begin{tabular}{clcc}
        \toprule
        \multicolumn{1}{c}{\textbf{Method}} & \multicolumn{1}{c}{\tiny\textbf{$N^{+}$:$N^{-}_{strong}$:$N^{-}_{weak}$}} & \multicolumn{1}{c}{\textbf{AE 2.0 (\%)}} & \multicolumn{1}{c}{\textbf{MT-Bench}} \\
        \midrule
        \multirow{2}{*}{Ablation} &\ \ \ \ \ \ \ \ 1:1:0 (DPO) & 17.43 / 15.24 & 7.55 \\ 
           & \ \ \ \ \ \ \ \ 1:0:2 & 9.37 / 6.74 & 6.57 \\ 
         \cdashlinelr{1-4}
          \multirow{3}{*}{EPA} & \ \ \ \ \ \ \ \ 1:1:1 & \textbf{21.14} / \textbf{20.55} & 7.29 \\
         & \ \ \ \ \ \ \ \ 1:1:2 & 19.20 / 19.26 & \textbf{7.71} \\
         & \ \ \ \ \ \ \ \ 1:1:6 & 16.63 / 15.78  & 7.57 \\
        \midrule
        Ablation & \ \ \ \ \ \ \ \ 1:3:0 (infoNCA) & 17.30 / 16.25 & 7.50\\
        \cdashlinelr{1-4}
          \multirow{4}{*}{EPA} & \ \ \ \ \ \ \ \ 1:3:2 & 22.03 / 21.44 & \textbf{7.58}  \\
          & \ \ \ \ \ \ \ \ 1:3:4 & 21.31 / 20.13 & \textbf{7.58}  \\
           & \ \ \ \ \ \ \ \ 1:3:6 & 24.01 / 23.44 & 7.35 \\
           & \ \ \ \ \ \ \ \ 1:3:8 & \textbf{24.54} / \textbf{23.75} & 7.19 \\
           & \ \ \ \ \ \ \ \ 1:3:10 & 23.58 / 22.78  & 7.43 \\
        \bottomrule
    \end{tabular}
    \caption{Ablation and variants of EPA with varying number of strong negatives ($N^{-}_{strong}$) and varying number of weak negatives ($N^{-}_{weak}$) in addition to the 1 positive ($N^{+}=1$) in the denominator of $\mathcal{L}^{\text{general}}_{\text{EPA}}$.}
    \label{tab:neg_ablation}
\end{table}
We also run ablation and variants of EPA for different numbers of strong and weak negatives. As shown in Table \ref{tab:neg_ablation}, we can see that although weak negatives are not competitive on its own, their presence together with strong negatives can greatly improve the alignment performance over alternatives with only strong negatives. Moreover, we also find that weak negatives can even improve DPO, but DPO is still worse than EPA in that scenario (Table \ref{tab:dpo_neg} in Appendix \ref{apdx:b}).
\subsubsection{EBM+tricks $>=$ BTM+tricks}

\begin{table}
    \centering
    \begin{tabular}{cccc}
        \toprule
        \multicolumn{1}{c}{\textbf{Pref}} & 
         \multicolumn{1}{c}{\textbf{Loss }} & \multicolumn{1}{c}{\multirow{2}{*}{\textbf{AE 2.0 (\%)}}} & \multicolumn{1}{c}{\textbf{MT-}} \\
        \textbf{Model} & \textbf{Modification} & & \textbf{Bench}\\
        \midrule
        \multirow{6}{*}{BTM} & N/A (DPO) & 17.43 / 15.24 & 7.55 \\
       \cdashlinelr{2-4}
       & $-ref+\mathcal{L}_{sft}$ (CPO)  & 13.63 / 10.34 & 7.11 \\
        \cdashlinelr{2-4}
       & $+{len}_{p}$ (R-DPO)& 19.10 / 16.71 & \underline{7.70} \\
        \cdashlinelr{2-4}
        & $-ref+{len}_{n}+m_{c}$ (SimPO) & 20.57 / 20.19 & 7.61 \\
        \cdashlinelr{2-4}
        & $+{w}_{op}$ (WPO)  & 21.90 / \underline{21.04} & 7.56 \\
        \cdashlinelr{2-4}
         & $+{m}_{c}$ & \underline{22.33} / 20.61 & 7.67 \\
        \midrule
       \multirow{3}{*}{EBM} & N/A (EPA) & 19.20 / 19.26 & \textbf{7.71}  \\
        \cdashlinelr{2-4}
        & $+{w}_{op}$ & 22.80 / 22.26 & 7.61 \\
        \cdashlinelr{2-4}
         & $+{m}_{c}$ & \textbf{23.00} / \textbf{22.47} & 7.68 \\
        \bottomrule
    \end{tabular}
    \caption{BTM vs. EBM with empirically effective loss modification tricks. 
    %We consider on-policy weighting ($+w_{op}$), length penalty ($+{len}_{p}$), length normalization ($+{len}_{n}$), constant margin ($+{m}_{c}$), removal of the reference model ($-ref$), and addition of a SFT loss ($+\mathcal{L}_{sft}$).
    }
    \label{tab:tricks}
\end{table}
\begin{figure}
  \includegraphics[width=0.42\textwidth]{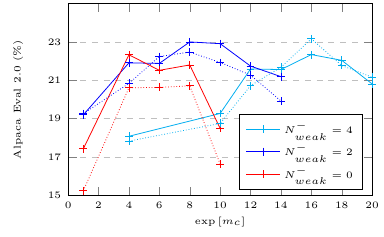}
  \caption{
  Performance of modified DPO ($N^{-}_{weak}=0$) and modified EPA ($N^{-}_{weak}>0$) with a margin $m_c$ added to $r_{\theta}(y_l|x)$. Solid lines represent the length-controlled win-rates, and dotted lines represent the raw win-rates.
  }
  \label{fig:margin}
\end{figure}
Since most loss modification tricks presented in the recent offline alignment literature are originally intended for BTM/DPO and do not necessarily make sense to EBM/EPA, we only consider two of them when applying to EPA. The first one is a constant margin $m_c$ added to the logit of $y_l$. The trick can be viewed as a loose numerical approximation to the general EPA where there are multiple $y_{l_k}$. For example, if $m_c=1.4$, we have $\exp{[r_{\theta}(x,y_l)+m_c]}=\exp{[m_c]}\times\exp{[r_{\theta}(x,y_l)]}\approx4\times\exp{[r_{\theta}(x,y_l)]}$. The second one is the on-policy weight proposed by \citet{zhou2024wpo}. It can be viewed as a curriculum learning technique which prioritizes samples that closely relate to the current policy distribution at each step.

As shown in Table \ref{tab:tricks}, we find that both $+w_{op}$ and $+{m}_{c}$ produce similar performance boost on EPA to DPO. Although the marginal boost on EPA is generally smaller than DPO, EPA with tricks is still better than DPO with tricks. However, the fact these tricks can still work on EPA also implies that there is still room for improvement. This may come from EPA not necessarily being the best algorithm to approximate our EBM's MLE.

We also study in detail how the value of ${m}_{c}$ influences DPO and EPA. As shown in Figure \ref{fig:margin}, we observe that a combination of higher ${m}_{c}$ and higher $N^{-}_{weak}$ tends to produce higher performance. A possible explanation for this is that as ${m}_{c}$ scales up the logit of the strong negative $y_l$ to loosely approximate the existence of multiple strong negatives, we get closer to the performance of the general EPA.

\section{Conclusion}
In this paper, we show both BTM and our EBM have the property that their MLE, if uniquely exists, is equivalent to the minimizer of the RLHF loss. However, the unique existence of EBM's MLE is guaranteed whereas that of BTM's MLE is not. This theoretical advantage implies that as long as the EBM's MLE is accurately found, we are bound to minimize the RLHF loss. But, the same claim does not hold for BTM. Although EPA is just an empirical attempt to approximate our EBM's MLE, it is already sufficient to outperform its counterpart -- DPO on open benchmarks, with or without loss modification tricks presented in previous work.
%In this paper, we point out that BTM's MLE is not guaranteed to exist uniquely. However, the MLE of our EBM (the Infinite Preference Model) has guaranteed unique existence. 
%This theoretical advantage manifests itself in empirical results where our proposal (EPA) to approximate the EBM's MLE outperforms its counterpart -- DPO on open benchmarks for preference alignment, with or without loss modification tricks presented in previous work. 

However, EPA is far from perfect. For example, relatively poorer computation and memory efficiency is a major handicap of EPA. Foreseeable future work includes finding better ways to perturb data or adopting more efficient methods to approximate the MLE. Loss modification tricks particularly tailored for EPA also remain to be explored.
\nocite{langley00}

\bibliography{reference}
\bibliographystyle{icml2025}

%%%%%%%%%%%%%%%%%%%%%%%%%%%%%%%%%%%%%%%%%%%%%%%%%%%%%%%%%%%%%%%%%%%%%%%%%%%%%%%
%%%%%%%%%%%%%%%%%%%%%%%%%%%%%%%%%%%%%%%%%%%%%%%%%%%%%%%%%%%%%%%%%%%%%%%%%%%%%%%
% APPENDIX
%%%%%%%%%%%%%%%%%%%%%%%%%%%%%%%%%%%%%%%%%%%%%%%%%%%%%%%%%%%%%%%%%%%%%%%%%%%%%%%
%%%%%%%%%%%%%%%%%%%%%%%%%%%%%%%%%%%%%%%%%%%%%%%%%%%%%%%%%%%%%%%%%%%%%%%%%%%%%%%
\newpage
\appendix
\section{Proof of the equivalence between slope-1 linearity and minimizer of the RLHF loss}
\label{apdx:a}
We do not claim any originality for the proofs given in this section because they are largely just paraphrased versions of the work by \citet{korbak2022reinforcement,rafailov2023direct} and others. We include them just for reference and completeness of the mathematical foundation shared by both DPO and EPA.
\label{apdx:linearity}
\begin{lemma}
\label{lemma:1}
The minimizer of the RLHF objective uniquely exists.
\end{lemma}

\begin{proof}
We show the minimizer $\pi_{r}$ can be analytically expressed by $\frac{1}{Z(x)}\pi_{\text{ref}}(y|x)\exp{\frac{1}{\beta}r(x,y)}$ where $Z(x)=\Sigma_{y}^{\infty}\pi_{\text{ref}}(y|x)\exp{\frac{1}{\beta}r(x,y)}$ (i.e., a normalizer to make $\pi_{r}$ a probabilistic distribution).

From the property of Gibb's inequality, we know:
\begin{equation}
\nonumber
  \begin{aligned}
    & \text{\small$\pi_{r}  = \frac{1}{Z(x)}\pi_{\text{ref}}(y|x)\exp{[\frac{1}{\beta}r(x,y)]}$}\\
      & \text{\small$\Leftrightarrow$}\\
      & \text{\small$\pi_{r} = \argminA_{\pi_{\theta}} \beta \text{KL}[\pi_{\theta}(y|x)||\frac{1}{Z(x)}\pi_{\text{ref}}(y|x)\exp{\frac{1}{\beta}r(x,y)}]$}
  \end{aligned}\hspace{-1.2em}
\end{equation}
We will complete the proof by showing the $\beta$KL-Divergence on the RHS of the above equation is the RLHF objective itself plus a constant w.r.t $\theta$:
\begin{equation}
\nonumber
  \begin{aligned}
    & \beta \text{KL}[\pi_{\theta}(y|x)||\frac{1}{Z(x)}\pi_{\text{ref}}(y|x)\exp{\frac{1}{\beta}r(x,y)}]\\
      & = \beta \mathop{\mathbb{E}}_{\pi_{\theta}(y|x)}[\log \frac{\pi_{\theta}(y|x)}{\frac{1}{Z(x)}\pi_{\text{ref}}(y|x)\exp{\frac{1}{\beta}r(x,y)}}]\\
      & = \beta \mathop{\mathbb{E}}_{\pi_{\theta}(y|x)}[\log \frac{Z(x)}{\exp [\frac{1}{\beta}r(x,y)]}+\log\frac{\pi_{\theta}(y|x)}{\pi_{\text{ref}}(y|x)}] \\
      & = \beta \mathop{\mathbb{E}}_{\pi_{\theta}(y|x)}[\log Z(x) - \frac{1}{\beta}r(x,y) +\log\frac{\pi_{\theta}(y|x)}{\pi_{\text{ref}}(y|x)}] \\
      & = -\mathop{\mathbb{E}}_{\pi_{\theta}(y|x)}[r(x,y)] + \beta \text{KL}[\pi_{\theta}(y|x)||\pi_{\text{ref}}(y|x)]\\
      &\ \ \ \ \ + \beta \log Z(x)\\
      & = \mathcal{L}_{RLHF}(\theta) + \beta \log Z(x)\\
  \end{aligned}\hspace{-1.2em}
\end{equation}
\end{proof}
\begin{definition}
We say a slope-1 linearity holds when:
$$ r_\theta(x, y) = r(x, y) + C(x) $$
where $r_\theta(x, y) = \beta \log \frac{\pi_{\theta}(y|x)}{\pi_{\text{ref}}(y|x)}$.
\end{definition}
\begin{theorem}[Theorem of necessity]
If $\pi_{\theta} = \pi_{r}$, then slope-1 linearity holds. 
\end{theorem}

\begin{proof}
    If $\pi_{\theta} = \pi_{r}$, then according to Lemma \ref{lemma:1}, we have:
    $$\pi_{\theta} = \frac{1}{Z(x)}\pi_{\text{ref}}(y|x)\exp{\frac{1}{\beta}r(x,y)}$$
    Take the logarithm of both sides of this equation, we have:
    $$\log \pi_{\theta} = \log \pi_{\text{ref}}+\frac{1}{\beta}r(x,y) - \log Z(x)$$
    After moving the two log terms to the same side, we get slope-1 linearity:
    $$\beta \log \frac{\pi_{\theta}}{\pi_{\text{ref}}} = r(x,y) - \beta\log Z(x)$$
\end{proof}
\begin{theorem}[Theorem of sufficiency]
If slope-1 linearity holds, then $\pi_{\theta} = \pi_{r}$. 
\end{theorem}

\begin{proof}
    From the Theorem of necessity, we know:
    $$\beta \log \frac{\pi_{r}}{\pi_{\text{ref}}} = r(x,y) - \beta\log Z(x)$$
    Substracting the slope-1 linearity from this equation, we get:
    $$\beta \log \frac{\pi_{r}}{\pi_{\text{ref}}} -\beta \log \frac{\pi_{\theta}}{\pi_{\text{ref}}} = - \beta\log Z(x) - C(x)$$
    Eliminating the non-zero $\beta$ from both sides and taking the exponential, we have:
    $$\frac{\pi_{r}}{\pi_{\theta}} = f(x)$$
    where $f(x) = \frac{1}{Z(x)\exp[\frac{1}{\beta}C(x)]}$.
    Moving $\pi_{\theta}$ to the RHS, we get:
    $$\pi_{r} = \pi_{\theta}f(x)$$
    Taking $\Sigma_{y}^{\infty}$ for both sides, we can sum up both distributions $\pi_{r}$ and $\pi_{\theta}$ to one:
    $$1=f(x)$$
    Therefore,
    $$\pi_{r} = \pi_{\theta}f(x) = \pi_{\theta}$$
\end{proof}
Note that from the above proof, we can easily get the following corollary because $f(x)=1\Leftrightarrow C(x)=-\beta\log Z(x)$.
\begin{corollary}
\label{cor1}
when a $r_{\theta}$ satisfies slope-1 linearity, it is unique.
\end{corollary}

\section{Theoretical Aspect of the Infinite Preference Model}
\label{apx:ipm}
We will first give our proof of the guaranteed unique existence of our IPM's MLE. Then, we will discuss how BTM is flawed for an infinite space of $y|x$.
\subsection{On IPM's MLE}
We make the following mild assumptions about the structure of human preference:
\begin{assumption}
\label{assume1}
    $\mathcal{D}_{\cdot|x}=\{y|p(y|x)>0\}$ is a finite set.
\end{assumption}
\begin{assumption}
\label{assume2}
    $r(x,y)\xrightarrow{}-\infty$ for any $y\notin \mathcal{D}_{\cdot|x}$ and $r(x,y)<+\infty$ for any $y\in \mathcal{D}_{\cdot|x}$.
\end{assumption}
Note that the above two assumptions are just one of many sufficient assumptions that make the partition function exist as a finite real number. Also note that the \emph{finity} of $\mathcal{D}_{\cdot|x}$ and the \emph{infinity} of the space of $y|x$ are two different things that can certainly co-exist. Namely, $\mathcal{D}_{\cdot|x}$ is a subset of the space of $y|x$. The number of $y$ outside of $\mathcal{D}_{\cdot|x}$ is still infinitely large. The two assumptions in plain words are simply that we assume humans will only possibly prefer a finite set of responses. Note that this does \emph{not} mean that the finite set $\mathcal{D}_{\cdot|x}$ cannot be very large.
\begin{definition}
The maximum likelihood estimation objective of IPM is the negative log-likelihood of preference data computed as follows:
$$ -\Sigma_{y}^{\infty} p(y|x)\log q_{\theta}(y|x)$$ 
where $p(y|x) = \frac{\exp [r(x,y)]}{\Sigma_{y^{'}}^{\infty}\exp [r(x,y^{'})]}$ and $q_{\theta}(y|x) = \frac{\exp [r_{\theta}(x,y)]}{\Sigma_{y^{'}}^{\infty}\exp [r_{\theta}(x,y^{'})]}$.
\end{definition}
Given the uniqueness in Corollary \ref{cor1}, we can argue the following:
\begin{theorem}[Theorem 3.1 in the main content of the paper]
\label{thrm:ipm}
The $r_{\theta}$ that satisfies the slope-1 linearity is the unique minimizer of the IPM's maximum likelihood estimation objective. 
\end{theorem}

\begin{proof}
Again, from the property of Gibb's inequality, we know:
\begin{equation}
\nonumber
  \begin{aligned}
    & q_{\theta}(y|x) = p(y|x)\\
      & \Leftrightarrow\\
      & q_{\theta}(y|x) = \argminA_{q} \text{KL}[p(y|x)||q(y|x)]
  \end{aligned}
\end{equation}
For the equation on the right, since $p(y|x)$ is a constant w.r.t $\theta$, we can find $q_{\theta}$ is the minimizer of IPM's objective:
\begin{equation}
\nonumber
  \begin{aligned}
    q_{\theta}(y|x) & = \argminA_{q} \text{KL}[p(y|x)||q(y|x)]\\
      & = \argminA_{q} \text{H}[p(y|x)]-\Sigma_{y}^{\infty} p(y|x)\log q(y|x)\\
      & = \argminA_{q} -\Sigma_{y}^{\infty} p(y|x)\log q(y|x)
  \end{aligned}
\end{equation}
We then show that $q_{\theta}(y|x) = p(y|x)$ is equivalent to slope-1 linearity to complete the proof by taking the logarithm of both sides:
\begin{equation}
\nonumber
  \begin{aligned}
     & r_{\theta}(x,y) - C_1(x) = r(x,y) - C_2(x)\\
      & \Leftrightarrow\\
      & r_{\theta}(x,y) = r(x,y) + C(x)
  \end{aligned}
\end{equation}
where $C_1(x)=\log\Sigma_{y^{'}}^{\infty}\exp [r_{\theta}(x,y^{'})]$, $C_2(x)=\log\Sigma_{y^{'}}^{\infty}\exp [r(x,y^{'})]$ and $C(x)=C_1(x)-C_2(x)$
\end{proof}
Note that similar proof does not apply to BTM. The fundamental reason is that the $C_1(x)$ and $C_2(x)$ only become constants when there is an infinity in the sum to cancel out all $y$.
\subsection{On Bradley-Terry Model's Flaw}
We will show in Proposition \ref{prop:btflaw} that a very likely choice of $p(y_w,y_l|x)$ will lead to multiple minimizers for the maximum likelihood estimation of BTM. There are also many other choices of $p(y_w, y_l|x)$ are known to cause the existence of multiple minimizers \cite{ford1957,pmlr-v162-bong22a}, such as when there is no 
 full connectivity of the graph made by pairs from $p(y_w, y_l|x)$, and when all $y$ candidates can only be paired with a single shared winning $\Tilde{y}$, etc. Therefore, there is no guarantee for the MLE's uniqueness without imposing additional constraints on $p(y_w,y_l|x)$ (i.e, how the pairs are sampled for DPO).  For the loosest \emph{sufficient} constraints discovered so far to ensure the uniqueness, one can refer to \citet{pmlr-v162-bong22a}. However, to the best of our knowledge, how such constraints can be applied to DPO has never been studied in the offline alignment literature, which is also out of the scope of this paper. Moreover, in the infinite-candidate scenario,  \textbf{a constraint that is \emph{both necessary and sufficient} for the uniqueness of BTM's MLE remains unknown} to this day. What makes BTM even more theoretically troublesome in the case of RLHF is that there is also an infinity for the space of $x$ as well. Therefore, strictly speaking, there is an infinite number of BTMs used in DPO. And, the $p(y_w,y_l|x)$ for every $x$ should ensure the uniqueness, in order to make DPO really work as expected. Interestingly, although our EPA loss also needs an infinite number of IPMs in the strict sense, Theorem \ref{thrm:ipm} (\ref{theorem:main}) ensures the MLE uniqueness of all the IPMs.
\begin{proposition}
\label{prop:btflaw}
If there exists a $y^*$ that will never be sampled (i.e., $p(y^*,\cdot|x)=0$ and $p(\cdot,y^*|x)=0$), then whenever there is a minimizer for Bradley-Terry's maximum likelihood estimation, it is not unique.
\end{proposition}

\begin{proof}
Without losing generality, we set $\beta=1$. 

Given the log ratio reward parameterization, we have an intrinsic constraint on $r_\theta$:
$$\sum\limits_{y'}^{\infty}\pi_{\text{ref}}\exp[r_{\theta}]=\sum\limits_{y'}^{\infty}\pi_{\theta}=1$$

If we assume that there is a unique minimizer $r_{\hat{\theta}}$ to BTM's maximum likelihood estimation, it certainly satisfies the above constraint:
$$\sum\limits_{y'}^{\infty}\pi_{\text{ref}}\exp[r_{\hat{\theta}}]=1$$

We will then show that another reward also follows the constraint (hence a valid log ratio reward) and shares the same expected data likelihood as $r_{\hat{\theta}}$, which contradicts the uniqueness of $r_{\hat{\theta}}$. We define the other reward as:
$$
\text{\footnotesize$\Tilde{r}_{\hat{\theta}}(x, y)=$}
\begin{cases}
    \text{\footnotesize$\log(\exp[r_{\hat{\theta}}(x, y) + A(x)]+\frac{1-exp[A(x)]}{\pi_{\text{ref}}}),$}& \text{\footnotesize$\text{if } y= y^{*}$}\\
    r_{\hat{\theta}}(x, y) + A(x),              & \text{\footnotesize otherwise}
\end{cases}
$$
where $A(x)$ can be any negative constant w.r.t $y$.
This reward satisfies the constraint because:

\begin{equation}
\nonumber
  \begin{aligned}
     &\sum\limits_{y'}^{\infty}\pi_{\text{ref}}\exp[\Tilde{r}_{\hat{\theta}}(x,y')]\\
      & =\sum\limits_{y'\neq y^*}^{\infty}\pi_{\text{ref}}\exp[\Tilde{r}_{\hat{\theta}}(x,y')]+\pi_{\text{ref}}\exp[\Tilde{r}_{\hat{\theta}}(x,y^*)]\\
      & =\sum\limits_{y'\neq y^*}^{\infty}\pi_{\text{ref}}\exp[r_{\hat{\theta}}(x,y')+A(x)]\\
      &\ \ \ \ \ \ +\pi_{\text{ref}}\exp[\Tilde{r}_{\hat{\theta}}(x,y^*)]\\
      & =\exp[A(x)]\sum\limits_{y'\neq y^*}^{\infty}\pi_{\text{ref}}\exp[r_{\hat{\theta}}(x,y')]\\
      &\ \ \ \ \ \ +\pi_{\text{ref}}\exp[\Tilde{r}_{\hat{\theta}}(x,y^*)]\\
      & =\exp[A(x)](1-\pi_{\text{ref}}\exp[r_{\hat{\theta}}(x,y^*)])\\
      &\ \ \ \ \ \ +\pi_{\text{ref}}\exp[\Tilde{r}_{\hat{\theta}}(x,y^*)] \\
      & =\exp[A(x)](1-\pi_{\text{ref}}\exp[r_{\hat{\theta}}(x,y^*)])\\
      &\ \ \ \ \ \ +\pi_{\text{ref}}\exp[r_{\hat{\theta}}(x,y^*) + A(x)] + (1-\exp[A(x)]) \\
      & =\exp[A(x)](1-\pi_{\text{ref}}\exp[r_{\hat{\theta}}(x,y^*)])\\
      &\ \ \ \ \ \ +\exp[A(x)](\pi_{\text{ref}}\exp[r_{\hat{\theta}}(x,y^*)] - 1) + 1 \\
      & =1
  \end{aligned}
\end{equation}
Note that the constraint makes $\Tilde{r}_{\hat{\theta}}$ correspond to a valid policy $\Tilde{\pi}_{\hat{\theta}}=\pi_{\text{ref}}\exp[\Tilde{r}_{\hat{\theta}}]$ that sums up to 1. The policy is also in the range of $[0, 1]$ because:

1) for $y=y^*$:
\begin{equation}
\nonumber
  \begin{aligned}
     &\Tilde{\pi}_{\hat{\theta}}=\pi_{\text{ref}}\exp[\Tilde{r}_{\hat{\theta}}]\\
      & = \pi_{\text{ref}}\exp[r_{\hat{\theta}}+A(x)]+1-\exp[A(x)]\\
    &=\exp[A(x)]\pi_{\text{ref}}\exp[r_{\hat{\theta}}]+1-\exp[A(x)]\\
    &=\exp[A(x)]\pi_{\text{ref}}\frac{\pi_{\hat{\theta}}}{\pi_{\text{ref}}}+1-\exp[A(x)]\\
    &=\exp[A(x)](\pi_{\hat{\theta}}-1)+1
  \end{aligned}
\end{equation}
\ \ \ \ \ \ \ and since $A(x)<0$ and $\pi_{\hat{\theta}}\in[0,1]$, we have:
\begin{equation}
\nonumber
  \begin{aligned}
     &1\geq\exp[A(x)](\pi_{\hat{\theta}}-1)+1\\
     &\geq \exp[0](\pi_{\hat{\theta}}-1)+1\\
     &=\pi_{\hat{\theta}}\geq 0
  \end{aligned}
\end{equation}
\ \ \ \ \ \ \ hence $\Tilde{\pi}_{\hat{\theta}}\in[0, 1]$;

2) for $y\neq y^*$:
\begin{equation}
\nonumber
  \begin{aligned}
     &\Tilde{\pi}_{\hat{\theta}}=\pi_{\text{ref}}\exp[\Tilde{r}_{\hat{\theta}}]\\
      & = \pi_{\text{ref}}\exp[r_{\hat{\theta}}+A(x)]\\
      &= \exp[A(x)]\pi_{\text{ref}}\exp[r_{\hat{\theta}}]
  \\
  &=\exp[A(x)]\pi_{\text{ref}}\frac{\pi_{\hat{\theta}}}{\pi_{\text{ref}}}\\
  &=\exp[A(x)]\pi_{\hat{\theta}}\\
  &\in[0, 1]
  \end{aligned}
\end{equation}

Then, we show that this valid log ratio reward is indeed another minimizer because it leads to the same expected likelihood of data as $r_{\hat{\theta}}$.
\begin{equation}
\nonumber
  \begin{aligned}
     &\mathop{\mathbb{E}}_{p(y_w,y_l|x)}\mathop{\mathbb{E}}_{p(y_w\succ y_l|x)}[\log \sigma (\Tilde{r}_{\hat{\theta}}(x, y_w) - \Tilde{r}_{\hat{\theta}}(x, y_l))]\\
      & = \sum\limits_{y_w,y_l}^{\infty}p(y_w,y_l|x)\mathop{\mathbb{E}}_{p(y_w\succ y_l|x)}[\log \sigma (\Tilde{r}_{\hat{\theta}}(x, y_w) - \Tilde{r}_{\hat{\theta}}(x, y_l))]\\
      & = \sum\limits_{y_w\neq y^{*},y_l\neq y^{*}}^{\infty}p(y_w,y_l|x)\cdot \\
      & \ \ \ \ \ \ \mathop{\mathbb{E}}_{p(y_w\succ y_l|x)}[\log \sigma (\Tilde{r}_{\hat{\theta}}(x, y_w) - \Tilde{r}_{\hat{\theta}}(x, y_l))] \\
      & = \sum\limits_{y_w\neq y^{*},y_l\neq y^{*}}^{\infty}p(y_w,y_l|x)\cdot \\
      & \ \ \ \ \ \ \mathop{\mathbb{E}}_{p(y_w\succ y_l|x)}[\log \sigma (r_{\hat{\theta}}(x, y_w) + A(x) - r_{\hat{\theta}}(x, y_l) - A(x))] \\
      & = \sum\limits_{y_w\neq y^{*},y_l\neq y^{*}}^{\infty}p(y_w,y_l|x)\cdot \\
      & \ \ \ \ \ \ \mathop{\mathbb{E}}_{p(y_w\succ y_l|x)}[\log \sigma (r_{\hat{\theta}}(x, y_w) - r_{\hat{\theta}}(x, y_l))] \\
      & = \sum\limits_{y_w,y_l}^{\infty}p(y_w,y_l|x)\mathop{\mathbb{E}}_{p(y_w\succ y_l|x)}[\log \sigma (r_{\hat{\theta}}(x, y_w) - r_{\hat{\theta}}(x, y_l))]\\
      & =\mathop{\mathbb{E}}_{p(y_w,y_l|x)}\mathop{\mathbb{E}}_{p(y_w\succ y_l|x)}[\log \sigma (r_{\hat{\theta}}(x, y_w) - r_{\hat{\theta}}(x, y_l))]
  \end{aligned}
\end{equation}
Finally, because the possible choices of $A(x)$ are infinite and by assuming the full representation capacity of a large neural network, $\Tilde{\pi}_{\hat{\theta}}$ can be represented by some other $\pi_{\Tilde{\theta}}$. 
\end{proof}
Note that if we are given the $p(y_w,y_l|x)$ defined in Proposition \ref{prop:btflaw}, even if we have an infinite amount of data, we still do not have a unique MLE. This refutes a popular claim that DPO will work better given enough data.
\subsection{Fitting IPM with Energy Discrepancy}
In our paper, since we simply adapt the general theorems given by \citet{schroder2023training} to fit the RLHF context, we only provide proof sketches for quick reference and claim no originality thereof. We encourage readers to refer to the original full proof if more details are needed.

\begin{proof}[Proof sketch for Theorem 3.2]
The first functional derivative of $ED_{x,p(y|x),p(z|y)}[E]$ (we use $E=-r$ to conform to the convention of EBMs) is given by:
\begin{equation}
\nonumber
  \begin{aligned}
     &\frac{d}{d\epsilon}ED_{x,p(y|x),p(z|y)}[E+\epsilon h]\\
      = & \mathop{\mathbb{E}}_{x,p(y|x)}[h(x,y)]-\mathop{\mathbb{E}}_{x,p(y|x),p(z|y)}\mathop{\mathbb{E}}_{p_{E,\epsilon}(y'|z)}[h(x,y')]
  \end{aligned}
\end{equation}
where
\begin{equation}
\nonumber
  \begin{aligned}
     p_{E,\epsilon}(y'|z)=\frac{p(z|y')\cdot \exp[-E(x,y')-\epsilon h(x,y')]}{\sum\limits_{y''}^{\infty}p(z|y'')\cdot \exp[-E(x,y'')-\epsilon h(x,y'')]}
  \end{aligned}
\end{equation}
Then, setting $\epsilon=0$ and $E=-r_{\text{true}}$, we get the first variation of $ED$ at $E=-r_{\text{true}}$ to be $0$ because the second term in the derivative becomes identical to the first term:
\begin{equation}
\nonumber
  \begin{aligned}
  &\mathop{\mathbb{E}}_{x,p(y|x),p(z|y)}\mathop{\mathbb{E}}_{p_{E=-r_{\text{true}},\epsilon=0}(y'|z)}[h(x,y')]\\
  &= \text{\small$\mathop{\mathbb{E}}_{x,p(y|x),p(z|y)}\sum\limits_{y'}^{\infty}\frac{p(z|y')\cdot \exp[r_{\text{true}}(x,y')]}{\sum\limits_{y''}^{\infty}p(z|y'')\cdot \exp[r_{\text{true}}(x,y'')]}\cdot h(x,y')$}\\
  &=\mathop{\mathbb{E}}_{x,p(y|x),p(z|y)}\sum\limits_{y'}^{\infty}\frac{p(z|y')p(y'|x)}{\sum\limits_{y''}^{\infty}p(z|y'')p(y''|x)}\cdot h(x,y')\\
  &=\mathbb{E}_{x}\sum\limits_{y}^{\infty}\sum\limits_{z}p(y|x)p(z|y)\frac{\sum\limits_{y'}^{\infty}p(z|y')p(y'|x)h(x,y')}{\sum\limits_{y''}^{\infty}p(z|y'')p(y''|x)} \\
  &=\mathbb{E}_{x}\sum\limits_{y'}^{\infty}p(y'|x)h(x,y')\sum\limits_{y}^{\infty}\sum\limits_{z}\frac{p(y|x)p(z|y)p(z|y')}{\sum\limits_{y''}^{\infty}p(z|y'')p(y''|x)}\\
  &=\mathbb{E}_{x}\sum\limits_{y'}^{\infty}p(y'|x)h(x,y')\sum\limits_{z}p(z|y')\frac{\sum\limits_{y}^{\infty}p(z|y)p(y|x)}{\sum\limits_{y''}^{\infty}p(z|y'')p(y''|x)} \\
  &=\mathbb{E}_{x}\sum\limits_{y'}^{\infty}p(y'|x)h(x,y')\sum\limits_{z}p(z|y') \\
  &=\mathbb{E}_{x}\sum\limits_{y'}^{\infty}p(y'|x)h(x,y') \\
  &=\mathbb{E}_{x,p(y|x)}h(x,y) \\
  \end{aligned}
\end{equation}
With the first variation being 0, we will then only need to show that the second variation of $ED$ at $E=-r_{\text{true}}$ is strictly positive to complete the proof that $E=-r_{\text{true}}$ is the global unique minimizer of $ED$. This can be done by showing that the second derivative becomes an expectation of $\text{Var}_{p(y|z)}[h(y)]$ which cannot be negative or zero because we assume $\text{Var}_{p(y|z)}[y]>0$ (i.e., $\text{Var}[Y|Z]>0$).
Concretely, the second derivative of $ED$ is given by:
\begin{equation}
\nonumber
  \begin{aligned}
     &\frac{d^2}{d\epsilon^2}ED_{x,p(y|x),p(z|y)}[E+\epsilon h]\\
      = &-\mathop{\mathbb{E}}_{x,p(y|x),p(z|y)}\frac{d^2}{d\epsilon^2}\mathop{\mathbb{E}}_{p_{E,\epsilon}(y'|z)}[h(x,y')]\\
      = &\text{\ (see \cite{schroder2023training}'s Lemma 2)}\\
      &\mathop{\mathbb{E}}_{x,p(y|x),p(z|y)}\left[\mathop{\mathbb{E}}_{p_{E,\epsilon}(y'|z)}[h^2(x,y')]-\big(\mathop{\mathbb{E}}_{p_{E,\epsilon}(y'|z)}[h(x,y')]\big)^2\right]\\
      = &\mathop{\mathbb{E}}_{x,p(y|x),p(z|y)}[\text{Var}_{p_{E,\epsilon}(y'|z)}[h(x,y')]]
  \end{aligned}
\end{equation}
Setting $\epsilon=0$ and $E=-r_{\text{true}}$, we will get the positive second variation of $ED$ at $E=-r_{\text{true}}$.
\end{proof}

\begin{proof}[Proof sketch for Theorem 3.3]
Given the property that $\Sigma_{y}p(z|y)f(y)=\Sigma_{y}p(y|z)f(y)$, the $\Sigma_{y'}p(z|y')$ in the definition of $ED$ becomes an expectation over $y$, i.e., $\mathop{\mathbb{E}}_{p(y'|z)}$. This enables a statistic estimate using the normal and a modified Strong Law of Large Numbers \cite{majerek2005conditional} for the expectations outside and inside of the logarithm, respectively.
\end{proof}

\subsection{The Perspective of Functional Analysis}
\begin{figure}
  \centering
  \includegraphics[width=0.48\textwidth]{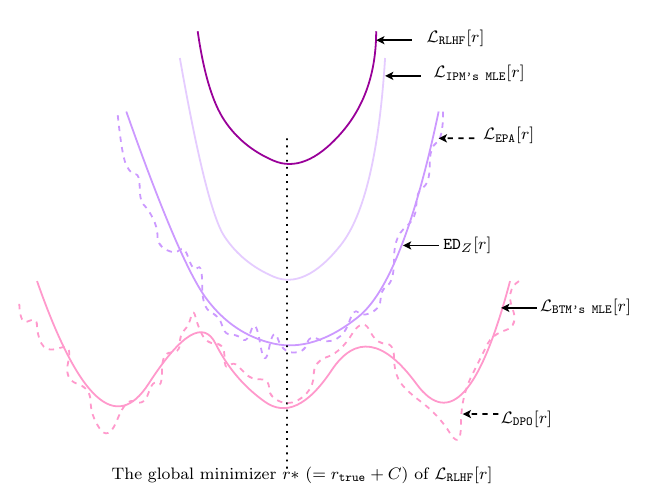}
  \caption{
  A simplified illustration of the topology of the functionals mentioned in the paper's theorems. The vertical axis represents the value of each functional. The horizontal axis represents the space of $r$.
  }
  \label{fig:functionals}
\end{figure}
An elegant way to understand the theoretical advantage of EBM/EPA over BTM/DPO is through the lens of functional analysis. As illustrated in Figure \ref{fig:functionals}, our EBM's maximum likelihood estimation loss, the energy discrepancy, and the RLHF loss can be regarded as three functionals sharing the same unique global minimizer. The EPA loss, being a statistical estimate\footnote{Strictly speaking, there is also a constant shift $\log M$, which we removed from the energy discrepancy to derive Eq.(10) because it has no topological impact in terms of optimization problems.} of the energy discrepancy, is yet another functional. However, the Strong Law of Large Numbers ensures that the error between it and the energy discrepancy will almost surely vanish with enough amount of negatives drawn from a properly designed data perturbing source $Z$. Therefore, one can view the EPA loss as a locally fuzzy approximation of the energy discrepancy. The maximum likelihood estimation loss of BTM is a functional that may have multiple minimizers. Therefore, without explicit constraints to prevent this from happening, chances are that optimizing the DPO loss (which is strictly speaking also just a statistical estimate of the maximum likelihood of data, probably with less fuzziness) will lead to a different solution than the minimizer of the RLHF loss. Increasing the amount of data for DPO can certainly mitigate the fuzziness but can do nothing to avoid the undesirable structure of BTM's maximum likelihood estimation loss governed by $p(y_w,y_l|x)$.
\section{More Experimental Details}
\label{apdx:b}
\subsection{Implementation Details}
\subsubsection{More hyperparameter details}
We use 8 A100/A800 GPUs (80G Memory) with ZeRO3 parallelism to train each model in this paper. Global batch size is fixed to 64. For experiments in Figure \ref{fig:kr_dynam}, we run two rounds with different seeds (0 and 1) for each model configuration. Other experiments are only conducted with seed 0. 
\subsubsection{Loss Modifications Tricks}
\begin{table}
    \centering
    \begin{tabular}{cc}
        \toprule
        \multicolumn{1}{c}{\textbf{Notation}} & 
        \multicolumn{1}{c}{\textbf{Trick function}} \\
        \midrule
        \multirow{2}{*}{$-ref$} & \multirow{2}{*}{$\log\frac{\pi_\theta(y|x)}{\pi_{\text{ref}}(y|x)}\xrightarrow{}\log \pi_\theta(y|x) $} \\
        & \\
       \cdashlinelr{1-2}
       \multirow{2}{*}{$+\mathcal{L}_{sft}$} & \multirow{2}{*}{$\mathcal{L}\xrightarrow{}\mathcal{L}-\log \pi_\theta(y_w|x) $} \\
       & \\
       \cdashlinelr{1-2}
       \multirow{2}{*}{$+{len}_p$} & \multirow{2}{*}{$r_\theta(y|x)\xrightarrow{}r_\theta(y|x)-\alpha|y|$} \\
       & \\
       \cdashlinelr{1-2}
       \multirow{2}{*}{$+{len}_n$} & \multirow{2}{*}{$r_\theta(y|x)\xrightarrow{}\frac{1}{|y|}r_\theta(y|x)$} \\
       & \\
       \cdashlinelr{1-2}
       \multirow{2}{*}{$+m_c$} & \multirow{2}{*}{$r_\theta(y_l|x)\xrightarrow{}r_\theta(y_l|x)+m_c$} \\
       & \\
       \cdashlinelr{1-2}
       \multirow{2}{*}{$+w_{op}$} & \multirow{2}{*}{$\mathcal{L}\xrightarrow{}\Pi_{*\in\{w,l\}}{(\Pi_{t}\frac{\pi_\theta(y_*^t|y_*^{0:t-1},x)}{\Sigma_{y'\in Voc}\pi^2_\theta(y^{'}|y_*^{0:t-1},x)})}^{\frac{1}{|y|}}\cdot\mathcal{L}$} \\
       & \\
        \bottomrule
    \end{tabular}
    \caption{Notation and function of loss modification tricks proposed in the offline alignment literature. We consider on-policy weighting ($+w_{op}$), length penalty ($+{len}_{p}$), length normalization ($+{len}_{n}$), constant margin ($+{m}_{c}$), removal of the reference model ($-ref$), and addition of a SFT loss ($+\mathcal{L}_{sft}$)}
    \label{tab:listoftricks}
\end{table}
In Table \ref{tab:listoftricks}, we list how the tricks are applied to the DPO loss or the narrow EPA loss. For more detailed properties of them, one can refer to the corresponding previous work. In summary, length penalty \cite{park2024disentangling}, length normalization \cite{yuan2023rrhf,hong2024orpo,meng2024simpo}, constant or dynamic margins \cite{meng2024simpo,amini2024direct}, removal of the reference model \cite{xu2024contrastive,hong2024orpo,meng2024simpo}, on-policy weighting \cite{zhou2024wpo} and addition of SFT loss \cite{xu2024contrastive,hong2024orpo}, etc.
\subsection{DPO vs. EPA from More Perspectives}

\subsubsection{Using weak negatives in DPO}
\begin{table}[ht!]
    \centering
    \begin{tabular}{clcc}
        \toprule
        \multicolumn{1}{c}{\textbf{Training Data}} & 
         \multicolumn{1}{c}{\textbf{Method}} & \multicolumn{1}{c}{\textbf{AE 2.0 (\%)}} & \multicolumn{1}{c}{\textbf{MT-Bench }} \\
        \midrule
       \multirow{3}{*}{UF-binarized}  & \ \ \ \ DPO & 17.43 / 15.24 & \underline{7.55} \\
       \cdashlinelr{2-4}
         & \ \ \ \ EPA (1:1:1) & \textbf{21.14} / \textbf{20.55} & 7.29 \\
         & \ \ \ \ EPA (1:1:2) & 19.20 / 19.26 & \textbf{7.71} \\
       \cdashlinelr{1-4}
       $+$ UF-weak$\times$1 & \ \ \ \ DPO & \underline{18.99} / 17.00 & 7.37 \\
       $+$ UF-weak$\times$2 & \ \ \ \ DPO & 18.49 / \underline{18.72}  & 7.42 \\
        \bottomrule
    \end{tabular}
    \caption{Adding the same number of weak negatives to pair-wise data for DPO does not show any advantage over EPA. $+$UF-weak$\times$1 means to add a copy of UF-binarized with the $y_l$ replaced by a random weak negative for each prompt. $+$UF-weak$\times$2 means to add 2 such copies.}
    \label{tab:dpo_neg}
\end{table}
It is also interesting to compare DPO and EPA with the computation complexity and number of weak samples as strict control variables. We implement this by directly using the same number of weak negatives as additional alternatives for $y_l$ in pair-wise data for DPO. However, we would like to emphasize that this is \emph{not} meaningful for refuting or supporting our core argument about our EBM's superiority over BTM as there is no obvious theoretical link between such usage and approximating the MLE of either our EBM or BTM. Suppressing the log ratio reward of weak negatives is just a necessary condition for the slope-1 linearity as opposed to a sufficient one. As shown in Table \ref{tab:dpo_neg}, we can see that there is no major advantage of such usage over EPA.
\subsubsection{Alignment tax}
\begin{table}
    \centering
    \begin{tabular}{lccc}
        \toprule
        \multicolumn{1}{c}{\textbf{Method}} & 
         \multicolumn{1}{c}{\textbf{GSM8k-5Shot}} & \multicolumn{1}{c}{\textbf{MMLU}} & \multicolumn{1}{c}{\textbf{Winograd}} \\
        \midrule
       SFT  & 0.421 (0.014)\ \  & \textbf{0.598} (0.004)\ \  & \textbf{0.800} (0.006) \\
       \cdashlinelr{1-4}
        +DPO & \textbf{0.463} (0.014)\ \  & 0.593 (0.004) & 0.790 (0.006) \\
        +EPA & 0.419 (0.014) & 0.591 (0.004) & 0.790 (0.006) \\
        \bottomrule
    \end{tabular}
    \caption{EPA has slightly higher alignment tax than DPO.}
    \label{tab:tax}
\end{table}
Preference alignment is usually associated with an alignment tax: forgetting certain capabilities (e.g., math problem solving) while enhancing others (e.g., safety, truthfulness, and helpfulness). We acknowledge that EPA might exhibit a higher alignment tax than DPO due to the results in Table \ref{tab:tax}, where we report metrics on GSM8k \cite{Cobbe2021}, MMLU \cite{Hendrycks2020}, and Winograd \cite{Tikhonov2021}. However, a more comprehensive future work on this issue is still necessary for a reliable conclusion. For example, although GSM8k and MMLU  are of relatively higher correlation with human preference than other Exact-Match-based benchmarks, they have poorer correlations than the ones used in the main paper (MT-Bench and Alpaca-Eval 2.0) \cite{dubois2024length}. Therefore, a more aggressive alignment method with human preference could cause the lower scores on GSM8k and MMLU.
%A similar phenomenon has also been previously reported for the usage of loss modification tricks such as SimPO.
% \subsubsection{More pretrained LLMs}
% To provide empirical evidence of the generality of our argument about the superiority of our EBM, we report in Table X the results of applying DPO and EPA on pretrained LLMs other than the Mistral model used in the main paper.
\subsubsection{Probing the slope-1 linearity}
Reward models in the past only need to satisfy ranking consistency with the corresponding true rewards. However, for ideal offline optimization of the RLHF loss, a learned log ratio reward and its corresponding true reward have to satisfy the slope-1 linearity, a much stronger requirement than ranking consistency. The reason is that, to satisfy ranking consistency, the relationship between the two rewards can be any monotonically increasing function, including those that are \emph{non-linear} or \emph{linear with the slope being any positive number}. 

Therefore, metrics that are based on ranking consistency to evaluate the log ratio reward (i.e., the learned reward model) is not meaningful when it comes to offline alignment. Instead, we evaluate the log ratio reward by probing how well the linearity is approximated. For this purpose, we need multiple samples of $y$ (responses) given an $x$ (prompt) and their true rewards. The test split of the Ultrafeedback data \cite{tunstall2023zephyr} can fulfill this purpose because there are four $y$ for each $x$ and they are scored using the same scoring scheme used for our training data (i.e., UF-all and UF-binarized). We randomly sample 500 prompts from the test split to speed up evaluation while preserving the general reliability.

\begin{table}
    \centering
    \begin{tabular}{lcc}
        \toprule
         \multicolumn{1}{c}{\multirow{3}{*}{\textbf{Method}}} & \multicolumn{1}{c}{\textbf{Pearson ($\uparrow$)}} & \multicolumn{1}{c}{\textbf{$\hat{\epsilon}$ ($\downarrow$) }} \\
         \cline{2-3}
          & slope-1: NA & slope-1: \checkmark\\
          & linearity: \checkmark & linearity: \checkmark\\
        \midrule
          \ \ \ \ DPO\ \ \ \  & 0.4693 & 5.78 \\
       \cdashlinelr{1-3}
          \ \ \ \ EPA (1:1:2)\ \ \ \  & \textbf{0.5808} & 5.26 \\
          \ \ \ \ EPA (1:3:2)\ \ \ \  & 0.5754 & \textbf{5.01} \\
        \bottomrule
    \end{tabular}
    \caption{Metrics to probe the slope-1 linearity. Both the average Pearson coefficient ($\in [-1, 1]$) and the average slope-1 linear regression error $\hat{\epsilon}$ show that EPA is closer to slope-1 linearity than DPO.}
    \label{tab:linear}
\end{table}

Firstly, we consider only how linear the relationship between the two rewards is, regardless of the slope. This is exactly the essence of Pearson correlation analysis. We compute the Pearson coefficient between the two rewards over the four $y$ for each prompt. Then, we compute the mean over all 500 prompts as a metric. We report the results in the first column of Table \ref{tab:linear}. 
%We also show the complete histogram distribution over all values of Pearson coefficient in Figure X. We can see that there are noticeably more prompts that have near-zero Pearson coefficients for DPO than EPA.

\begin{figure}[ht]

  \includegraphics[width=0.48\textwidth]{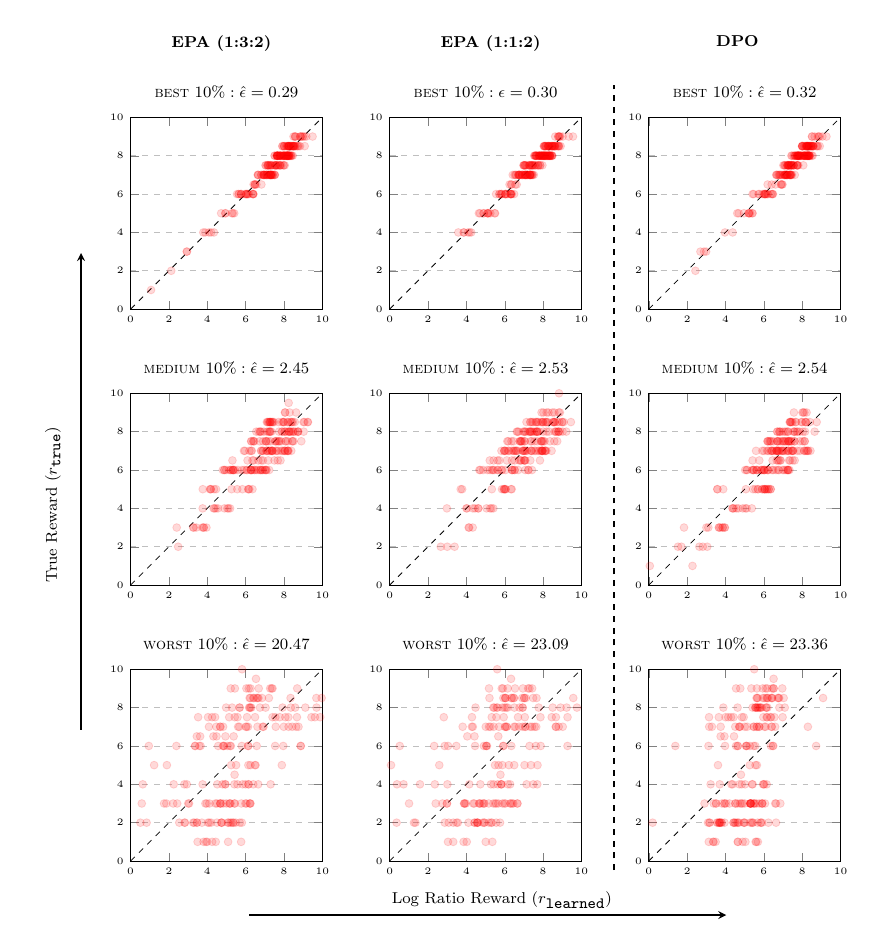}
  \caption{
  EPA vs DPO in terms of the slope-1 linearity. Each average slope-1 linear regression error $\hat{\epsilon}$ is computed on 500 $\times$ 10\% $=$ 50 prompts. Although the difference between EPA and DPO only becomes noticeable for the ``WORST 10\%" group in the visualization (i.e., how close the partially transparent red dots are to the diagonal), the difference in $\hat{\epsilon}$ is conspicuous.
  }
  \label{fig:linearity}
\end{figure}

Secondly, we study both the ``slope-1" and the ``linearity". We do this via linear regression with the slope fixed to 1. Specifically, given a prompt, we need to fit a linear regression model $r_{\text{learned}} = 1\cdot r_{\text{true}} + b$ to the 4 coordinates of $(r_{\text{learned}}, r_{\text{true}})$. With simple algebra, the optimal value of $b$ that minimize the linear regression error $\epsilon=\Sigma|r_{\text{true}}-r_{\text{learned}} + b|^2$ can be analytically expressed as $\hat{b}=(1/4)\cdot\Sigma(r_{\text{learned}} - r_{\text{true}})$. Thus, we use $\hat{b}$ to compute the minimal error for each given prompt, and then compute the average minimal error $\Hat{\epsilon}$ over all 500 prompts as the metric (see the second column of Table \ref{tab:linear}). We can also shift the four $r_{\text{learned}}$ for each prompt by the constant $-\hat{b}$, which should move all regression lines for the slope-1 linearity to the same location: the diagonal $r_{\text{learned}} = 1\cdot r_{\text{true}}$. This allows us to visualize the overall degree of how well the slope-1 linearity is approximated. As shown in Figure \ref{fig:linearity}, we can see that for the top 10\% prompts with the best $\hat{\epsilon}$ (i.e., smallest $\hat{\epsilon}$), the slope-1 linearity is well approximated for both EPA and DPO. However, we can observe that DPO is slightly off from the slope-1 linearity for the medium 10\% and much so for the worst 10\%. On the other hand, for EPA, although the points are also gradually spreading out when we move towards the worst 10\%, they are still distributed along the direction of the diagonal. This phenomenon means that EPA is closer to the slope-1 linearity than DPO, especially for the worst group of prompts.
% \onecolumn

% You can have as much text here as you want. The main body must be at most $8$ pages long.
% For the final version, one more page can be added.
% If you want, you can use an appendix like this one.  

% The $\mathtt{\backslash onecolumn}$ command above can be kept in place if you prefer a one-column appendix, or can be removed if you prefer a two-column appendix.  Apart from this possible change, the style (font size, spacing, margins, page numbering, etc.) should be kept the same as the main body.
%%%%%%%%%%%%%%%%%%%%%%%%%%%%%%%%%%%%%%%%%%%%%%%%%%%%%%%%%%%%%%%%%%%%%%%%%%%%%%%
%%%%%%%%%%%%%%%%%%%%%%%%%%%%%%%%%%%%%%%%%%%%%%%%%%%%%%%%%%%%%%%%%%%%%%%%%%%%%%%

\end{document}